\documentclass{article}

\PassOptionsToPackage{numbers, compress}{natbib}
\usepackage[main,final]{neurips_2026}
\usepackage{graphicx}
\usepackage{enumitem}

\usepackage[utf8]{inputenc} 
\usepackage[T1]{fontenc}    
\usepackage{hyperref}       
\usepackage{url}            
\usepackage{booktabs}       
\usepackage{amsfonts}       
\usepackage{nicefrac}       
\usepackage{microtype}      
\usepackage{xcolor}         
\usepackage{amsmath}
\usepackage{caption}  

\begin{document}
\title{Learning Options for Compositional \\ Motor Control with Adapter Banks}

%

\author{ Sreejan Kumar\textsuperscript{1,2} \quad
Marcelo G. Mattar\textsuperscript{2} \thanks{equally contributing senior authors}\footnotemark[1] \quad
Lea Duncker\textsuperscript{1}\footnotemark[1] \\[6pt]
\textsuperscript{1}Columbia University \quad
\textsuperscript{2}New York University \quad
}

\maketitle

\begin{abstract}
Learning flexible motor primitives is a hallmark of skilled motor control. Recent neuroscience theory proposes that motor primitives may be implemented as low-rank perturbations of a shared recurrent network, but leaves open how such a system is learned. We translate this principle into a novel architecture for learning motor skills end-to-end: a shared recurrent core modulated by a bank of residual adapters, each selected by a discrete latent code. Trained on closed-loop biomechanical control, the adapters develop emergent low-rank perturbations of the recurrent dynamics despite no architectural rank constraint, placing task representations in disparate subspaces of the shared core network. A simple high-level policy over the learned options, optimized while the whole network is frozen, sequences the low-rank adapters to produce novel out-of-distribution movements. We demonstrate the ability to generalize to novel motor sequences within the closed-loop control setting, improving on the generalization error of a task-input-conditioned multitask baseline by approximately sevenfold.
\end{abstract}

\section{Introduction}
In the movie \emph{The Karate Kid}, how did Mr. Miyagi train his student to win a fighting tournament through seemingly unrelated chores like waxing cars and painting fences? The capacity to generate flexible behavior from a finite set of reusable primitives that are recombined and rescaled across new contexts is a hallmark of skilled motor control in humans \citep{flash2005motor, heald2021contextual}. Reproducing this capacity in artificial agents has been a central goal of hierarchical and modular control: how can a learning system discover reusable building blocks of behavior, and flexibly recombine them to solve new problems \citep{khetarpal2022towards, duncker2020organizing}?

To answer this question, we first ask: what form should these reusable building blocks take? In hierarchical reinforcement learning, reusable primitives are commonly formed as \emph{options}: temporally extended actions, each with its own internal low-level policy, invoked by a high-level policy that decides which option to engage at each moment. Methods for discovering such skills learn both their individual low-level actions as well as how they can be recruited by the high-level policy \citep{eysenbach2018diversity,kipf2019compile,bacon2017option}. In the brain, however, different movements draw on shared cortical circuitry \citep{churchland2012neural}, raising the question of how distinct, reusable operations can be implemented within a common network. One possibility is that a motor primitive corresponds to a modulation of that network's ongoing dynamics. Selecting a primitive would then configure how the network transforms its current state and sensory feedback into movement, allowing the same operation to act under different conditions.

Cortico--basal ganglia--thalamo--cortical circuits provide a candidate architecture for implementing this proposal. In the framework of \citet{logiaco2021thalamic}, cortex supplies shared recurrent dynamics, while thalamocortical loops modify its effective connectivity. The basal ganglia acts as a selector of thalamic channels to determine which modifications are engaged, allowing distinct dynamical motifs to be recruited and sequenced within the same cortical network. This framework establishes how a bank of thalamic modulatory signals can generate composable motor operations. It leaves open, however, how the cortical network, the modulations acting upon it, and the control programs selecting them can be jointly learned to produce compositional motor control.

We address this learning problem with CABRA (Compositional Adapter Banks for Recurrent Architectures), a shared recurrent cortical network modulated by a bank of learned residual adapters representing thalamocortical loops, whose recruitment is determined by a basal ganglia controller. CABRA learns through complementary observation and execution passes. During observation, the cortical network processes an expert demonstration and supplies activity to the controller, which infers a sequence of adapters. During execution, the selected adapters modify cortical dynamics to generate behavior in closed loop, and execution error trains their effects. This couples learning which operations to recruit with learning how those operations act within a shared network. We implemented CABRA on the closed-loop musculoskeletal control setting of \citet{lazzari2025multitasking}. Three properties emerge from training. First, despite each adapter being initialized as a full-rank residual modulator, the adapters learn to apply emergent \emph{low-rank} perturbations to the core network, recovering the structural assumption of \citet{logiaco2021thalamic}. Second, task representations are factored into disparate subspaces of a shared recurrent substrate, factoring the compositional structure of the task suite. Third, the trained adapter library supports out-of-distribution \emph{compositional transfer}: a soft policy over the learned adapters, optimized while the rest of the network is frozen, sequences existing primitives into a novel trajectory the network was never trained on, outperforming a task-input-conditioned multitask baseline by approximately sevenfold on a novel two-petal movement. Together, these results show how learning within a cortico--basal ganglia--thalamic architecture can establish reusable motor operations that can provide the foundation to support more complex behavior over the same computational repertoire.

\section{Low-rank Adapters via Thalamocortical Architectures in Neuroscience}
\begin{figure}[h]
\centering
\includegraphics[width=\textwidth]{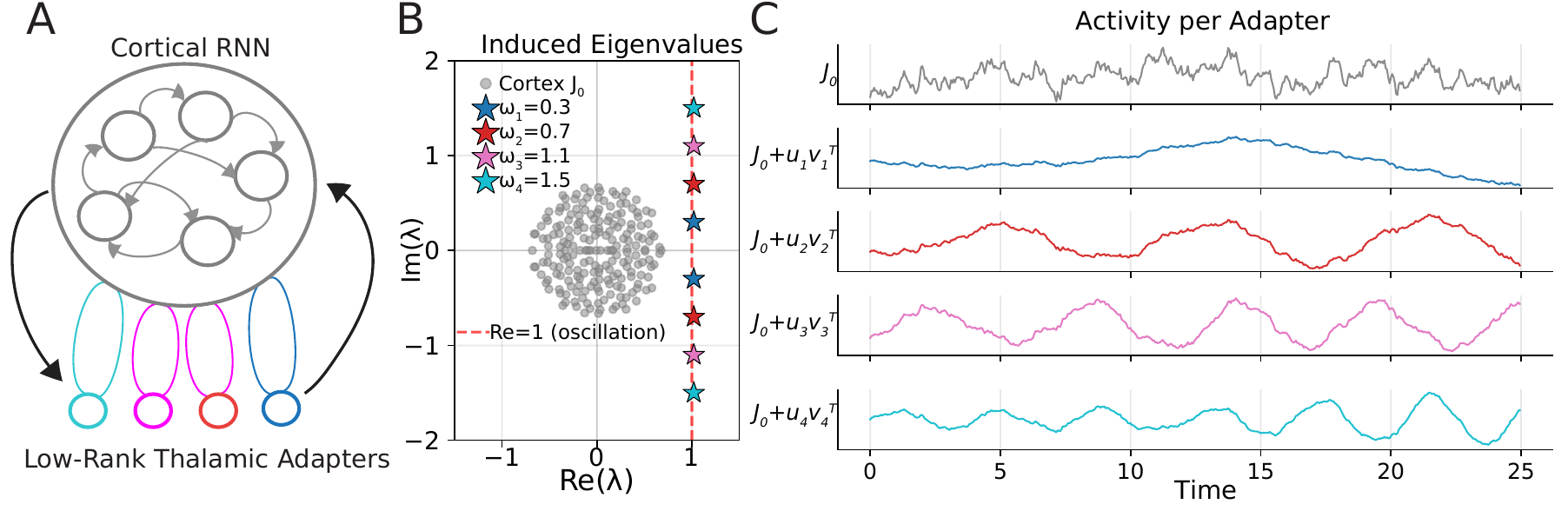}
\caption{\textbf{Adapter Banks as Thalamocortical Loops}
\textbf{(A)} Schematic of the framework of \citet{logiaco2021thalamic}: thalamic inputs $\mathbf{u}_k$ drive a cortical RNN with recurrence $\mathbf{J}_0$. A rank-one perturbation $\mathbf{u}_k \mathbf{v}_k^\top$ can induce oscillatory dynamics at a target frequency.
\textbf{(B)} Eigenvalue placement. Gray: eigenvalues of the bare cortex $\mathbf{J}_0$. Stars: target locations on the oscillation line (Re$(\lambda) = 1$) for four motifs at different frequencies $\omega_k$. Each low-rank perturbation shifts a conjugate pair of eigenvalues to the corresponding target. \textbf{(C)} Cortical activity projected onto a one-dimensional readout. Each motif $J_0+u_kv_k^{T}$ sustains a different oscillation at its target frequency $\omega_k$, demonstrating that rank-one perturbations selectively recruit the cortical network to produce frequency-specific dynamics. }
\label{fig:logiaco}
\end{figure}

Theoretical work has proposed an architecture for composing motor primitives into flexible sequences, inspired by the connectivity between motor cortex, thalamus and basal ganglia in the brain \citep{logiaco2021thalamic}. Here, motor primitives are implemented through an interplay of the recurrence in motor cortex, and cortex-thalamus-cortex projections (thalamocortical loops, Figure \ref{fig:logiaco}A). Distinct thalamic units participate in different thalamocortical loops, and activations of subsets of thalamic units via basal ganglia inputs can thus act as a low-rank perturbation of the effective recurrent dynamics in the motor cortex.
%
This allows the system to generate complex movements by sequencing low-rank perturbations of the same shared cortical substrate. 

\label{sec:logiaco}
%
%

Let $\mathbf{h} \in \mathbb{R}^N$ denote the cortical state and $\mathbf{J}_0 \in \mathbb{R}^{N \times N}$ the recurrent connectivity in cortex. Each thalamocortical channel $k$ consists of a corticothalamic readout $\mathbf{v}_k \in \mathbb{R}^N$ and a thalamocortical projection $\mathbf{u}_k \in \mathbb{R}^N$. The cortical dynamics evolve as
\begin{equation}
    \tau \dot{\mathbf{h}} = -\mathbf{h} + \mathbf{J}_0 \mathbf{h} + \sum_k \mathbf{u}_k \mathbf{v}_k^\top \mathbf{h} = -\mathbf{h} + \left(\mathbf{J}_0 + \sum_k \mathbf{u}_k \mathbf{v}_k^\top \right) \mathbf{h}.
\end{equation}
The thalamocortical loop therefore acts as a low-rank perturbation to the effective cortical dynamics. In the linear setting, \citet{logiaco2021thalamic} showed that this low-rank perturbation can be tuned to precisely modulate the eigenvalues of the effective dynamics (Figure~\ref{fig:logiaco}B). It can therefore change the behavior of the network (e.g. oscillation frequency of activity patterns, Figure~\ref{fig:logiaco}C) and introduce particular \emph{dynamical motifs} that would not be expressible by $\mathbf{J}_0$ alone.
%
%
This hence provides a mechanism for compositional motor control: the activity patterns needed to generate diverse movements can be constructed from a recurrent network by sequentially engaging different thalamocortical channels.

The framework in \citet{logiaco2021thalamic} showed how to design readout vectors $\mathbf{v}_k$ to achieve a particular change in eigenvalue of the effective dynamics and proposed that the selection mechanism can be implemented via disinhibition from basal ganglia circuits. However, they did not address how such an architecture could be \emph{learned} more generally, and what mechanisms might govern motif \emph{selection}.
In the following section, we build on the conceptual framework of \citet{logiaco2021thalamic} to arrive at an algorithm for the learning and selection of motifs in more general, nonlinear settings. 




\section{Learning Adapters for Flexible Motor Control}
\label{sec:tc-real-world}

We propose CABRA (Compositional Adapter Banks for Recurrent Architectures) as a framework for learning both which operations to recruit and what those operations do. We implement this through two complementary uses of a shared recurrent network: observing a demonstration to infer a control program, and acting under that program to reproduce the behavior (Figure~\ref{fig:architecture}B,C). These modes are paired within each training update.

\subsection{Motor task and demonstrations}
We adopt the closed-loop biomechanical control task from \citet{lazzari2025multitasking}. In this task, an RNN is required to control a two-link biomechanical arm with six muscles to generate a variety of movement types: straight reaches, clockwise and counterclockwise circular arcs, and sinusoidal trajectories. The RNN receives proprioceptive feedback (hand position, muscle length, muscle velocity) at each timestep. Movements come in two variants: \emph{half} (start $\to$ target) and \emph{full} (start $\to$ target $\to$ start). This yields ten distinct task conditions.

We use the model from \citet{lazzari2025multitasking} as a teacher network that can generate expert trajectories. The teacher network receives a one-hot \emph{rule input} at every timestep, specifying which task is currently active (Figure~\ref{fig:architecture}A), alongside speed, go-cue, and target information, as well as proprioceptive feedback. As was demonstrated in \citet{lazzari2025multitasking}, RNNs develop shared dynamical structure across movement types after training, allowing for reuse of the same computations across the different tasks. 
While this solution is compositional, the rule input supplies the network a particular organization of composable units rather than discovering them from learning. In the next section, we explore an alternative architecture with a bank of learnable adapters that can be flexibly selected to compose movements.

\subsection{CABRA: Compositional Adapter Banks for Recurrent Architectures}
\begin{figure}[h]
    \centering
    \includegraphics[width=\textwidth]{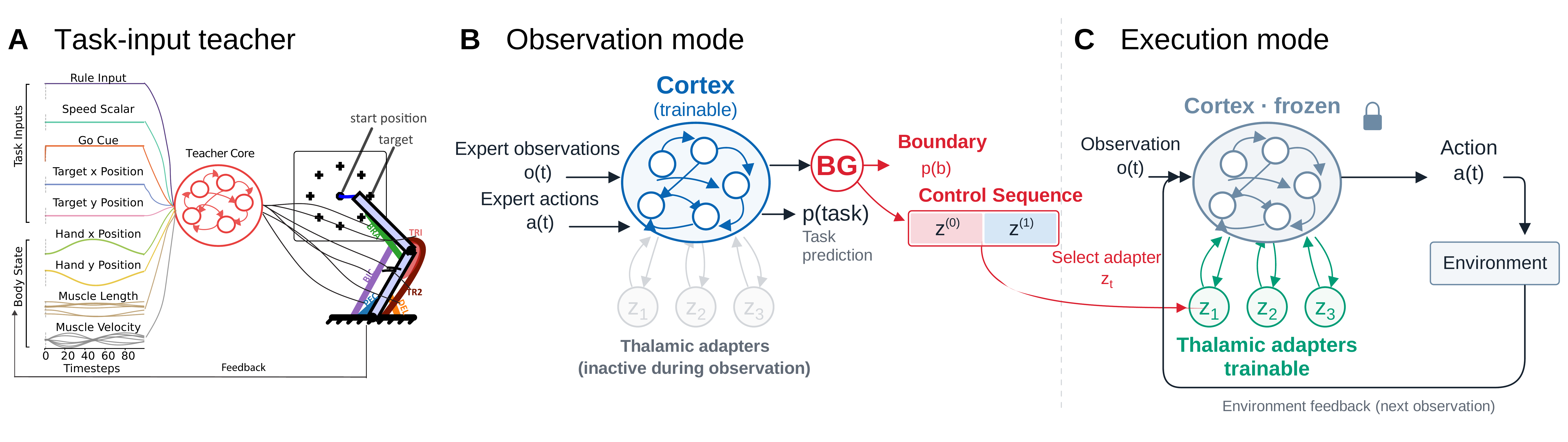}
    \caption{\textbf{An imitation learning framework to train flexible motor adapters}
    \textbf{(A)} We train a standard task-input-conditioned recurrent network controlling a two-link, six-muscle arm to supply expert demonstrations \citep{lazzari2025multitasking}.
    \textbf{(B)} In observation mode, CABRA's cortical network processes expert observations and actions without adapter modulation (note that CABRA's observations do not contain the rule input). Its activity supports a basal-ganglia-like (BG) selector that infers two adapter identities and their transition boundary, as well as an auxiliary task-identity prediction.
    \textbf{(C)} In execution mode, the inferred sequence recruits thalamic adapters that modulate the same cortical network to generate actions and receive environmental feedback. Execution error trains the adapters and readout. }
    \label{fig:architecture}
\end{figure}
To instantiate a brain-inspired principle for flexible motor control, we develop a general architecture which learns to observe expert demonstrations, segment them into learned discrete latent skills \citep{kipf2019compile}, and then use this decomposition to execute the same movement by selecting adapters to modulate the shared recurrent core. CABRA (Compositional Adapter Banks for Recurrent Architectures) comprises a shared recurrent core, a selector, and a bank of residual adapters. Their functional roles correspond to cortex, basal ganglia, and thalamocortical loops, respectively. The cortical core is a shared LSTM. The BG controller comprises readouts that predict adapter identities and transition boundaries from cortical activity. Each adapter is a learned nonlinear residual transformation of that activity, allowing the selected operation to depend on the current cortical state.

\paragraph{Observation mode: inferring what to recruit.}
During observation, the shared cortical network processes the expert’s sensory observations and muscle commands, without the rule input and with adapters inactive (Figure~\ref{fig:architecture}B). An auxiliary readout predicts the demonstrated task from cortical activity pooled over the first segment. This objective encourages representations of the identity of the observed behavior, motivated by evidence that cortical regions represent the goals of observed actions across changes in their immediate movement trajectories \citep{hamilton2007motor}. The BG controller reads cortical activity to infer a control program using the segmentation framework of CompILE \citep{kipf2019compile}. We prescribe two segments, corresponding to extension and return, while learning their transition time and adapter identities. For each segment, the controller predicts a boundary distribution $p(b)$ over timesteps and a categorical distribution $p(z)$ over adapters. The complete program is inferred from the expert demonstration before execution. 

During training, the code distribution is sampled with the straight-through Gumbel-softmax estimator, yielding a one-hot sample $z_{m,k} \in \{0,1\}$ with $\sum_k z_{m,k}=1$ in the forward pass so that exactly one adapter is active per segment. The boundary distribution is sampled with Gumbel-softmax at low temperature, inducing a soft segment assignment $\sigma_m(t) \in [0,1]$ that concentrates near $\{0,1\}$ except in the narrow window around the segment transition. At evaluation, both samples become hard via argmax. Half of the codes ($z_0, \ldots, z_4$) are reserved for segment 0 (extension); the other half ($z_5, \ldots, z_9$) for segment 1 (retraction). This phase split is implemented through hard masking of invalid logits and is the only structural prior we impose on the code identities themselves. The assignment of tasks to specific extension and retraction codes, and whether tasks reuse codes altogether, is up to the BG controller to learn. 

\paragraph{Execution mode: learning the recruited operations.} During execution, the inferred program recruits adapters that modulate the shared cortical network as it controls its own arm through sensory feedback (Figure~\ref{fig:architecture}C). The network receives neither expert muscle commands nor the rule input. Execution error directly optimizes the adapters, muscle readout, and execution-input projection, allowing the selected operations to acquire the effects needed for accurate behavior. This allocation is motivated by evidence that motor thalamus provides ongoing input essential to cortical movement dynamics, together with recent evidence that cerebellar-recipient motor thalamic neurons encode positional errors during reaching \citep{tien2025neurons,sauerbrei2020cortical}. These findings support a role for thalamocortical interactions in the ongoing regulation of movement.

The cortical recurrent weights are held fixed within the execution rollout and receive no direct execution-error gradient through that recurrence. However, the execution error can still shape the cortical recurrent weights indirectly through the BG controller's selection of the control program. Because adapter choices and their timing depend on cortical activity during observation, improving those choices also trains the cortical representations from which they are inferred. This constitutes controller-mediated cortical learning, related to proposals that basal ganglia selection guides the development of cortical representations and associations \citep{pasupathy2005different,ashby2007neurobiological}.

We impose no explicit rank constraint on the adapters. Each adapter (one per latent code) $\Delta_k: \mathbb{R}^{N} \rightarrow \mathbb{R}^{N}$ is a residual modulator of the core network's hidden state and any low-rank structure they develop in training emerges from the optimization itself, a property we examine quantitatively in Section~\ref{sec:results}. At each decoder timestep $t$, the core recurrent network produces a hidden state $h_t$ from observations and the previous state, and the adapter perturbs it residually. Let $z^{(0)},z^{(1)}$ denote the codes sampled by the encoders for both segments respectively, and let $\sigma(t)=\sum_{\tau \leq t}b_\tau$ denote the cumulative probability the segment boundary has occured by time $t$. At decoder timestep $t$, the core network produces a hidden state $h_t$ and the active adapters perturb it residually:
\begin{equation}
\tilde{h}_t=h_{t}+(1-\sigma(t))\Delta_{z^{(0)}}(h_t)+\sigma(t)\Delta_{z^{(1)}}(h_t)
\label{eq:adapter-update}
\end{equation}
Because $\sigma(t)$ is sharply concentrated around the sampled boundary, the perturbation collapses to $\Delta_{z^{(0)}}(h_t)$ during the extension segment and $\Delta_{z^{(1)}}(h_t)$ during the retraction segment, with brief soft mixing only across the transition. The muscle readout reads from $\tilde{h}_t$, the arm advances one-step through differentiable physics, and $\tilde{h}_t$ is fed back to the core network as the next-step recurrent input. The decoder receives the same task inputs and closed-loop body-state feedback the teacher uses during training, but never the rule input. Removing the adapters or replacing them with a single shared modulator prevents the network from effectively learning (Supplementary Figure~\ref{fig:adapter-ablation}).

\paragraph{Training objectives.}
We train the full system end-to-end with a hand-trajectory $L_1$ loss against the teacher's trajectories, KL regularization on boundary and code distributions, and standard activity and muscle-effort regularizers following \citet{lazzari2025multitasking}. Two auxiliary losses improve training stability and sharpen code routing: a behavioral discriminator that predicts the active code from segment-pooled muscle commands (DIAYN, \citealt{eysenbach2018diversity}), which reads only the six-dimensional muscle output and predicts the code that produced it; and an $L_1$ loss between student and teacher muscle excitations. Neither auxiliary loss directly determines the assignment of tasks to codes: the discriminator operates on muscle output rather than routing, and excitation matching shapes muscle commands rather than code logits. We show in Supplementary Figure~\ref{fig:auxloss-ablation} that both auxiliary losses can be ablated without losing the architectural transfer learning advantage.

We train CABRA only on compound trajectories (FullReach, FullCircleClk, FullCircleCClk, Figure-8, Figure-8 Inv) and never on the half-task primitives that compose them. Any decomposition into reusable half extension and retraction primitives must therefore arise from training rather than from being explicitly demonstrated as a half movement.

\section{Results}
\label{sec:results} 


\subsection{Compound performance and meaningful code routing}

\begin{figure}[h]
    \centering
    \includegraphics[width=\textwidth]{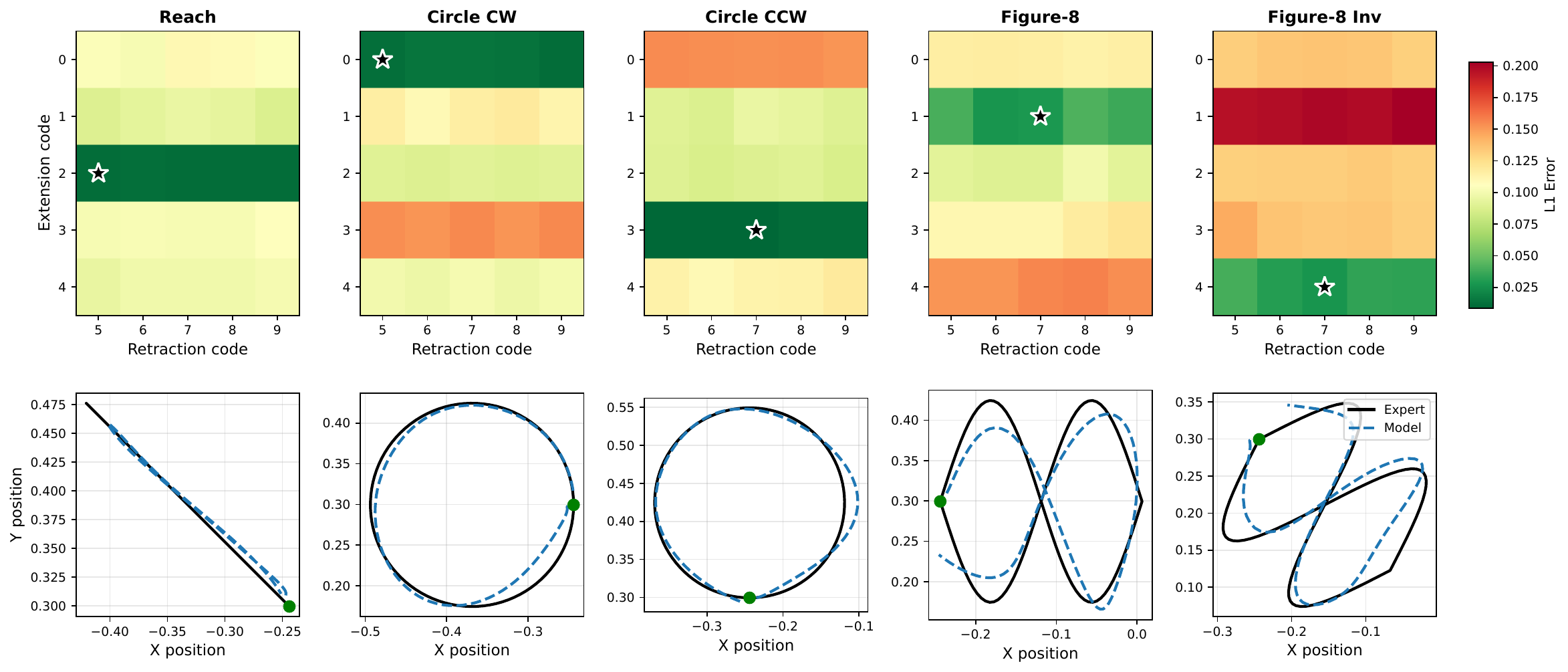}
    \caption{\textbf{CABRA execution of compound movements.}
    \emph{Top row:} Hand-trajectory L1 error of the trained CABRA under all 25 forced (extension code, retraction code) pairs, for each of the five compound tasks. Rows index the five extension codes ($z_0$--$z_4$); columns index the five retraction codes ($z_5$--$z_9$). Stars mark the natural code pair selected by the encoder.
    \emph{Bottom row:} Expert (black) and CABRA (blue, dashed) hand trajectories under the encoder's natural code choice. Green dots mark the start position.}
    \label{fig:behavior}
\end{figure}

We first verify that CABRA successfully reproduces the teacher's compound movements (Figure~\ref{fig:behavior}, bottom row). Trained only on compound trajectories and without ever being supplied the rule as input, CABRA traces the expert circles, figure-eights, and reaches with end-to-end hand L1 of approximately order $10^{-2}$~m on all five tasks. The reach is essentially exact; the figure-eights show closed-loop wobble deviations, but recover the general shape.

The more revealing analysis is the forced-code sweep (Figure~\ref{fig:behavior}, top row). For each compound task, we run the trained decoder under all 25 possible (extension code, retraction code) combinations and measure hand-trajectory error against the expert. If the discovered codes index meaningful primitives, forcing the model to use the wrong code should degrade behavior. If instead the codes are interchangeable, the heatmaps should be roughly uniform.

The pattern in the extension dimension shows specialized codes. Each task's natural extension code lies in a distinct row: Reach selects $z_2$, Circle CW selects $z_0$, Circle CCW selects $z_3$, Figure-8 selects $z_1$, and Figure-8 Inverse selects $z_4$. Forcing any other extension code raises error substantially. All five extension codes are populated, with a one-to-one assignment from tasks to codes. The training signal carries task-discriminative information (the encoder predicts task identity through an auxiliary readout, and the muscle output is matched to the teacher's task-specific commands) but does not specify the assignment from tasks to codes. The network is free to consolidate multiple tasks onto a single code or to populate the available codes one-to-one, and we observe the latter.

The retraction dimension is partially shared rather than one-to-one: Reach and Circle CW both select $z_5$ while Circle CCW, Figure-8, and Figure-8 Inverse all select $z_7$. Within a task's preferred extension code, swapping the retraction code from $z_5$ to $z_9$ produces only modest changes in trajectory error. This asymmetry between extensions and retractions tracks from the fact that the task prediction of the encoder is read out from the midpoint of the trial (specifically, the task readout is applied only to the first-segment encoding) and the kinematics of the task suite (several compound retractions converge to similar end-state trajectories).

\subsection{Emergent Low-Rank Structure}
\label{sec:geometry}

\begin{figure}[h]
    \centering
    \includegraphics[width=0.7\textwidth]{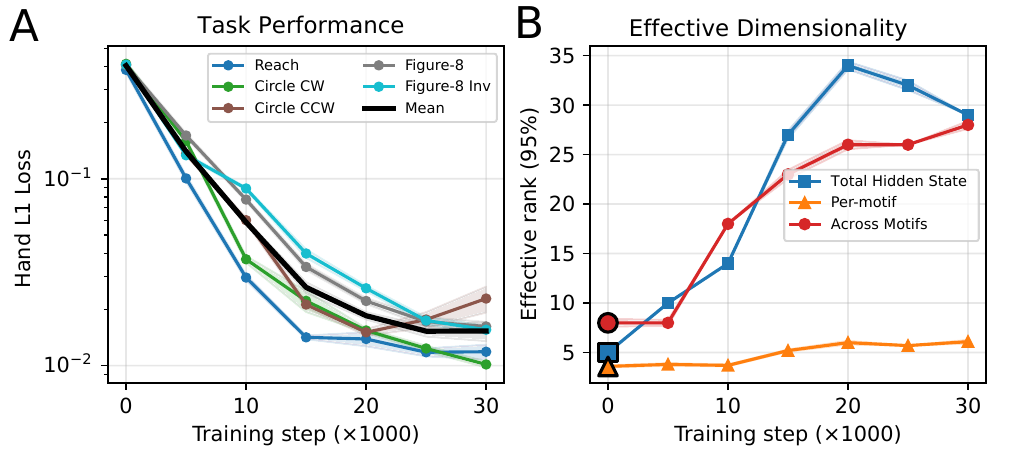}
    \caption{\textbf{Emergent low-rank structure in the adapter bank.}
    \emph{(A)} Held-out hand $L_1$ per task and mean over training. \emph{(B)} Effective rank (95\% variance) of three quantities computed from closed-loop rollouts: the recurrent core's hidden-state manifold, the per-adapter perturbation subspace (averaged across adapters), and the union of all adapter perturbations. Bands show bootstrap uncertainty (standard deviation across resampled rank estimates); the individual-adapter curve averages adapter ranks. These are activity dimensions, not weight-matrix ranks.}
    \label{fig:linear}
\end{figure}

We now ask whether the trained architecture exhibits the particular structural assumption that was made within \citet{logiaco2021thalamic}: motifs that are individually low-dimensional and jointly span the cortical state space. Recall that we impose no rank constraint on the motif adapters, because each is initialized as a full-rank residual modulator with an unconstrained $H \to H$ map. Any low-rank structure they develop in training is therefore a property the architecture discovers, not one we imposed.

Figure~\ref{fig:linear}A shows that all five compound tasks learn simultaneously. We characterize the dimensionality of the trained adapter bank's perturbation space relative to the dimensionality of the recurrent core's hidden-state manifold (Figure~\ref{fig:linear}B). Under the core-focused analysis protocol (Supplementary Methods), the recurrent core's hidden-state activity reaches an effective rank of $\sim 30$ by end of training (out of the available $H = 256$ dimensions), indicating the core develops a structured low-dimensional manifold. Each individual adapter's perturbation subspace converges to an effective rank of only $\sim 5$--$6$, which is well below both the core activity dimensionality and the available adapter dimensionality. And the union of all ten adapters' perturbations spans an effective rank of $\sim 28$, nearly matching the core's full hidden-state manifold. Here, dimensionality refers to the expressed perturbations over sampled states, not the algebraic rank of adapter weight matrices. The distinctive result is the contrast between individually low-dimensional perturbations and their richer combined span within a learned recurrent network. Individual adapter perturbations remained lower-dimensional than core activity with smaller banks of six or eight adapters (Supplementary Figure~\ref{fig:adapter-capacity}).

\subsection{Shared core, separated representations}
\label{sec:dynamics-subspaces}
\begin{figure}[h]
\centering
\begin{minipage}{0.58\textwidth}
\centering
\includegraphics[width=\textwidth]{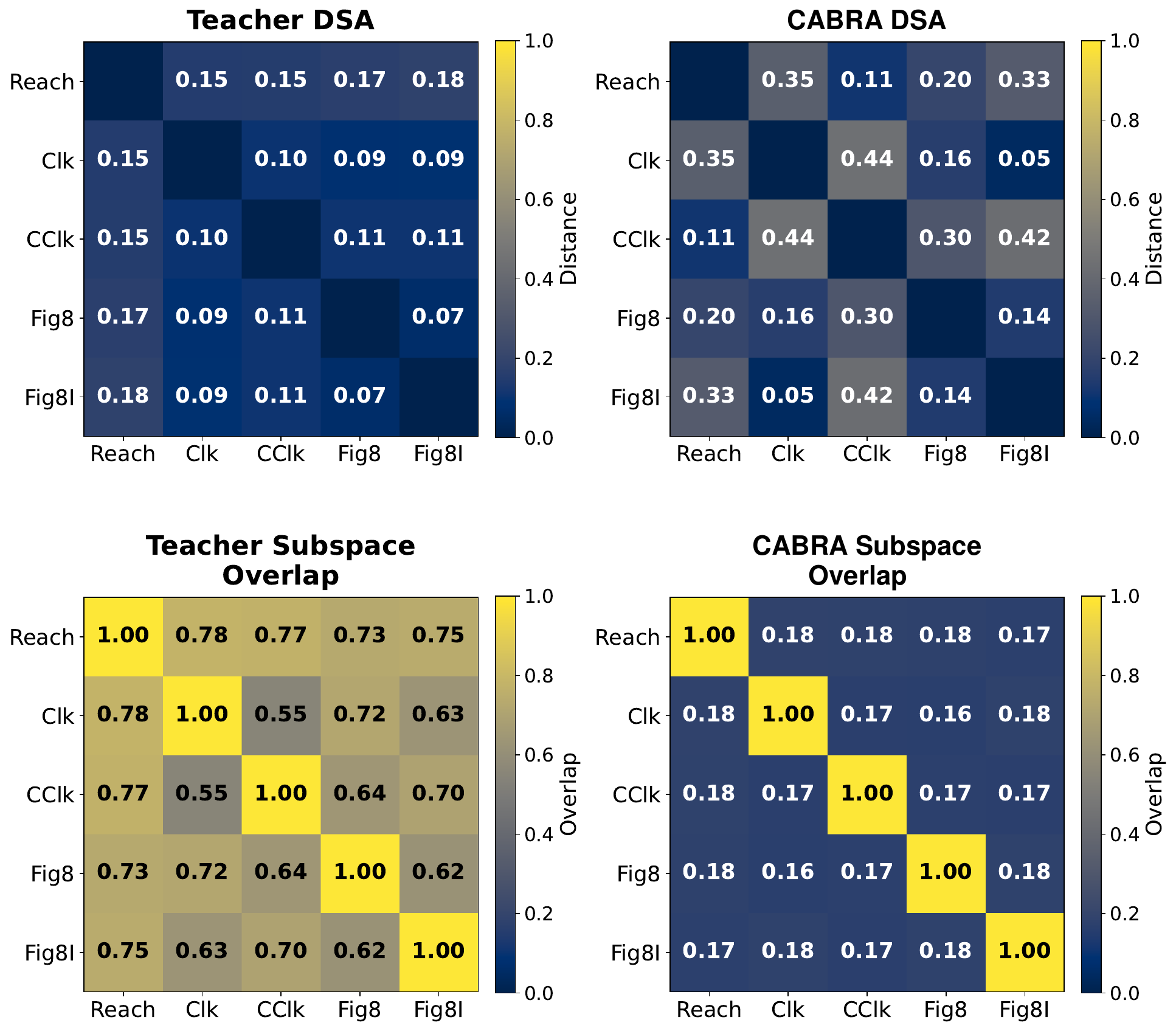}
\end{minipage}\hfill
\begin{minipage}{0.4\textwidth}
\caption{\textbf{Task-input baseline versus CABRA dynamics and representational geometry.} Pairwise dynamical similarity (DSA distance, top row; lower means more similar dynamics) and principal-subspace overlap (bottom row; mean squared cosine of principal angles between the dominant cortical activity subspaces, higher means more shared geometry) between the five compound tasks, computed for the teacher (left) and CABRA (right).}
\label{fig:dsa-subspaces}
\end{minipage}
\end{figure}

We now compare the dynamical and representational organization of CABRA to the teacher, a vanilla closed-loop multitask rule input-conditioned RNN. The two networks were trained on the same compound movements, although our network was only trained on compound movements while the teacher's training also included half extension-only movements. The teacher is a recurrent network that is modulated by a one-hot rule input, while CABRA is also a recurrent network but is modulated by learned adapters. We characterize each network along two complementary axes: \emph{representational geometry} (how recurrent hidden state activity for different tasks occupies the state space, measured via principal-subspace overlap) and \emph{dynamical structure} (how the underlying recurrent dynamics differ across tasks, measured via DSA). Figure~\ref{fig:dsa-subspaces} shows that this architectural difference produces a sharp dissociation along the representational axis, with corresponding structure along the dynamical axis.

\textbf{Representational geometry.} Following \citet{lazzari2025multitasking}, we compute the principal-subspace overlap between pairs of tasks via principal-angle analysis. For each task $A$, we collect recurrent hidden state activity over the trajectory into a matrix $\mathbf{X}_A \in \mathbb{R}^{N \times T}$ ($N=256$ units, $T$ timesteps) and extract the top $m$ principal components $\mathbf{P}_A \in \mathbb{R}^{N \times m}$ explaining $\geq 95\%$ of task variance. We use $m_{\text{teacher}}=12$ and $m_{\text{CABRA}}=25$, each matching its network's typical effective rank under the $95\%$ variance threshold. The pairwise principal-subspace overlap between tasks $A$ and $B$ is then
$ \frac{1}{m}\sum_{i=1}^{m}\cos^{2}\theta_{i}$ where $ \{\cos\theta_i\}_{i=1}^{m} = \mathrm{svd}\!\left(\mathbf{P}_A^{\top}\mathbf{P}_B\right)
$
and $\theta_i$ are the principal angles between the two subspaces. 
The teacher's pairwise overlap (Figure~\ref{fig:dsa-subspaces}, bottom-left) is uniformly high, with off-diagonal mean $0.69$. Task representations live in nearly the same recurrent subspace, consistent with the shared-manifold organization reported by \citet{lazzari2025multitasking}. CABRA's overlap (bottom-right) is uniformly low, with off-diagonal mean $0.18$, less than $2\times$ its random baseline. Task representations in CABRA are highly separated in state space. This is the core empirical realization of the framework from \citet{logiaco2021thalamic}: a shared substrate modulated by discrete perturbations into minimally-overlapping subspaces. Removing the phase mask, so that every adapter could be selected for either extension or return, preserved both low-dimensional adapter perturbations and separated task representations (Supplementary Figure~\ref{fig:no-phase-mask}).

\textbf{Dynamical structure.} We assess the similarity of recurrent dynamics across tasks using Dynamical Similarity Analysis (DSA, \citealt{ostrow2023beyond}), again following the protocol of \citet{lazzari2025multitasking}. 
The teacher's DSA matrix (Figure~\ref{fig:dsa-subspaces}, top-left) is uniformly low, with off-diagonal mean $0.12$: dynamics are similar across all task pairs. This is the within-manifold dynamical reuse that produces the teacher's single shared subspace. CABRA's DSA matrix (top-right) tells a different story. Off-diagonal mean is $0.25$, materially higher than the teacher's, but the values are not uniform in that they are organized along behavioral axes. Pairs sharing rotational sense have low DSA distance (Circle CW vs.\ Figure-8 Inverse: $0.05$; Circle CCW vs.\ Reach: $0.11$), while rotation-opposite pairs separate sharply (Circle CW vs.\ Circle CCW: $0.44$; Circle CCW vs.\ Figure-8 Inverse: $0.42$). The adapters do not produce small residual perturbations of a strictly common dynamical system, but rather genuinely distinct dynamical systems organized by behavioral kinematics.

These two findings are complementary. The teacher reuses both dynamics and representations: its compositional structure lives in a single shared subspace, modulated by explicit rule inputs. CABRA factors the problem differently. Its recurrent \emph{substrate} is shared: a single LSTM, the same recurrence, the same input projection, and the same readout. However, the adapter perturbations introduce structured dynamical differences and place task representations in highly separated subspaces. Where the teacher's compositional structure is tangled within a single manifold, CABRA's is geometrically factored.

\subsection{Compositional transfer to a novel two-petal trajectory}
\label{sec:transfer}

\begin{figure}[h]
    \centering
    \includegraphics[width=\textwidth]{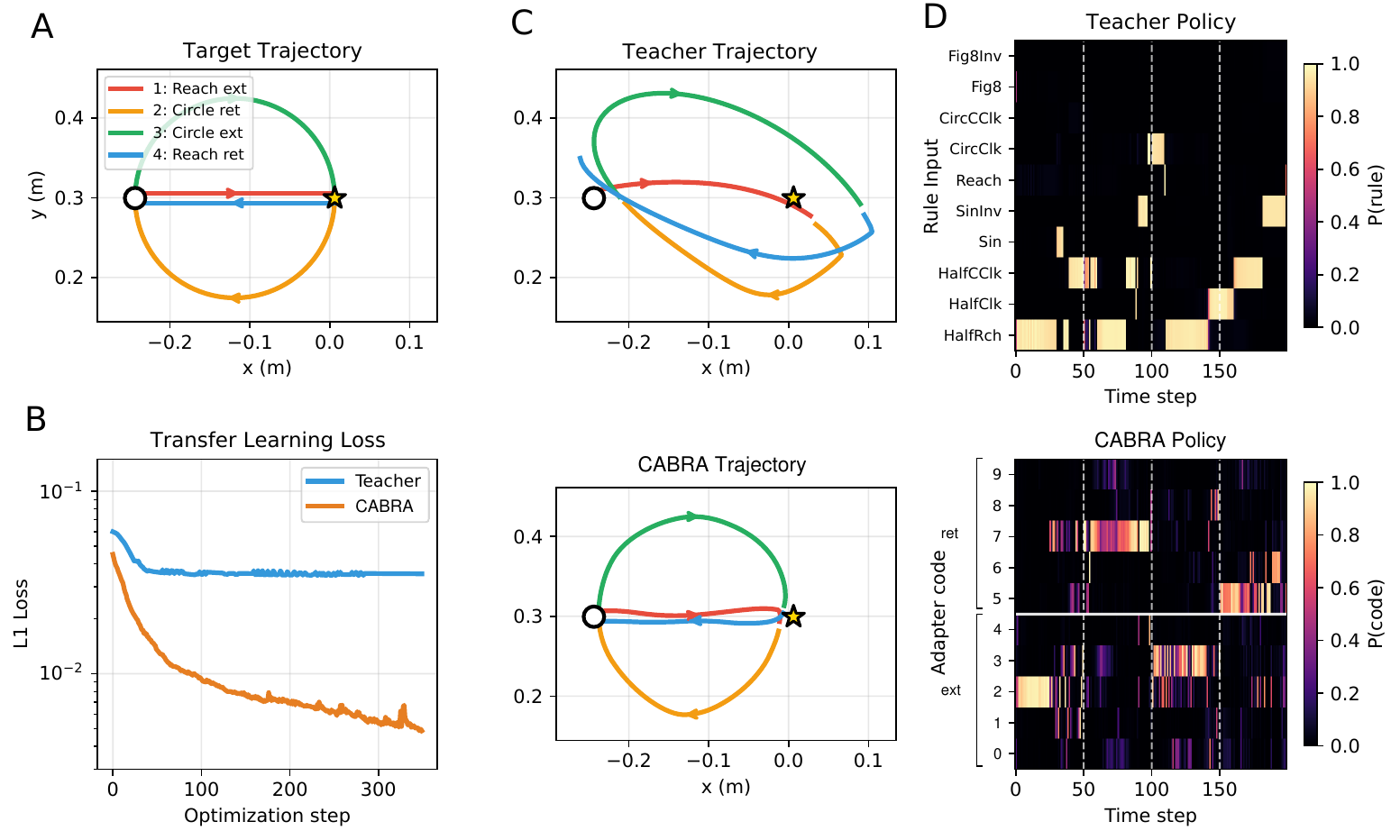}
    \caption{\textbf{Out-of-distribution compositional transfer.}
    \emph{(A)} The two-petal target trajectory, composed of four movement segments stitched together in a novel combination.
    \emph{(B)} Hand $L_1$ during transfer optimization, in which both networks are frozen and a 10-dimensional soft policy over the network's task interface is optimized for 350 steps via gradient descent. Final teacher $L_1 \approx 0.04$~; final CABRA $L_1 \approx 0.005$~.
    \emph{(C)} Final teacher (top) and CABRA (bottom) hand trajectories under the optimized policies. Open circle marks the home position; star marks the reach target.
    \emph{(D)} Final policies. Top: teacher rule-input weights over time. Bottom: CABRA adapter-code weights. Adapter IDs are consistent across rows; brackets mark the extension and return training pools, while transfer optimization can recruit the full bank.}
    \label{fig:transfer}
\end{figure}

The previous section established that CABRA's task representations occupy highly separable subspaces while the teacher's collapse into a single shared manifold. If this geometric factoring is what enables compositional reuse, then mixing primitives to compose a novel trajectory should be substantially easier in CABRA. We test this prediction directly.

We construct a synthetic four-segment trajectory (Figure~\ref{fig:transfer}A) that splices primitive movements in pairings neither network was trained on. Segment 1 is a straight extension followed by Segment 2's curved retraction, which contains a kinematic discontinuity at the target where the network must transition from a straight-line motor program to a curved-arc one. Segment 3 is a curved extension followed by Segment 4's straight retraction and represents the same discontinuity in the opposite direction. Neither (straight ext, curved ret) nor (curved ext, straight ret) appeared in either network's training data. We freeze both networks and optimize a $10$-dimensional soft policy over the network's task interface (only learning a $T \times10$ matrix): rule-input weights for the teacher and adapter-code weights for CABRA. We use gradient descent on the hand-trajectory $L_1$ to the two-petal target (Figure~\ref{fig:transfer}B). Both networks use the same optimizer, the same target, the same initialization heuristic, and the same evaluation procedure. The only thing that differs is the frozen substrate.

CABRA tracks the target to within $L_1=0.005$ while the teacher plateaus at $L_1\sim 0.04$, approximately sevenfold higher error (Figure~\ref{fig:transfer}B,C). CABRA's optimized policy (Figure~\ref{fig:transfer}D, bottom) shows that mixing concentrates around the segment transitions, but extends into within-segment timesteps as well (most prominently during Segment 3). CABRA also produces qualitatively distinct retractions across the two compositions: $z_7$ (producing a curved retraction) follows the straight Segment-1 extension, while $z_5$ (straight) follows the curved Segment-3 extension. Although forced-code analysis (Figure~\ref{fig:behavior}) showed that the choice of retraction code mattered little for in-distribution compound performance, in this OOD setting CABRA's selection policy meaningfully differentiates retraction primitives to match the trajectory's demands. The teacher's policy (Figure~\ref{fig:transfer}D, top), in principle, could mix multiple rule inputs to compose the target (and a few timesteps show such co-activation), but the transfer optimization consequently gravitates toward solutions where one rule dominates at a time, leaving the teacher unable to manage transitions within novel splice-compositions. The architectural advantage holds on a second more difficult out-of-distribution shape (Supplementary Figure~\ref{fig:butterfly}). CABRA's advantage is preserved when both networks are restricted to hard discrete selection at each timestep ($2.5\times$ gap in best error, $2.2\times$ over the final 50 updates; Supplementary
Figure~\ref{fig:transfer-hard}).

We also evaluated CABRA on D4RL Kitchen, a standard benchmark used in hierarchical reinforcement learning in which a 3D robotic arm completes household manipulation tasks, where a reward-trained policy over frozen adapters achieved higher early-training reward and comparable late-training performance to the skill-learning method SPiRL \citep{pertsch2021accelerating} (Supplementary Figure~\ref{fig:kitchen-rl}).

\section{Discussion}
\label{sec:discussion}
We investigated how reusable motor operations can be learned within a shared recurrent network whose dynamics are selectively modulated. CABRA couples learning to infer control programs from demonstrations with learning the effects of the operations those programs recruit. In closed-loop musculoskeletal control, this procedure establishes a repertoire that supports both trained movements and an untrained movement acquired through new control over the frozen network. These results extend a proposed circuit mechanism for composing motor motifs into a more general learning architecture, showing how operations acquired in particular behavioral contexts can remain useful beyond those contexts.

An important extension of the framework of \citet{logiaco2021thalamic} is that shared cortical dynamics and the primitives that recruit them are learned together. Their theory showed how structured thalamocortical loops can configure a fixed, randomly connected recurrent network to generate distinct motor motifs. In CABRA, the recurrent network itself acquires structure during learning, motivated by the rich movement-generating dynamics observed in motor cortex \citep{churchland2012neural}. Learning therefore shapes both the dynamical resources shared across movements and the adapter operations that selectively recruit them for individual movements. The resulting movements occupy differentiated cortical subspaces while retaining shared dynamical structure organized by behavioral kinematics (Figure~\ref{fig:dsa-subspaces}). Distinct motor primitives can thus draw on common learned dynamics while remaining separately accessible to control. Their separation does not require independent recurrent networks, and their sharing does not require identical cortical activity. This organization emerges as the shared substrate and its selectively recruited modulations develop together.

Related approaches offer complementary accounts of how shared computation can support flexible behavior. \citet{costacurta2024structured} configure recurrent computation by modulating gains on a learned low-rank connectivity basis. CABRA instead learns nonlinear residual operations around a full-width recurrent core, with low-dimensional perturbations emerging through training rather than being imposed architecturally. Work by \citet{shan2025separating} and \citet{bakermans2025compositional} likewise emphasizes how established computational structure can support new tasks through changes in control or inference. CABRA operates on a similar principle, but learns this structure instead. The differentiated task subspaces also connect to \citet{duncker2020organizing}, who showed how the organization of recurrent dynamics into orthogonal subspaces can mitigate interference during sequential continual learning. Our transfer experiment addresses a complementary question: whether new behavior can be acquired by changing control while preserving the learned repertoire. Although differentiated representations and successful reuse coexist here, establishing which features of that organization cause transfer would require more targeted interventions.

This account also links two perspectives on basal ganglia function: selecting actions \citep{mink1996basal} and influencing the detailed kinematics of learned movements \citep{dhawale2021basal}. When the selected object is an operation on ongoing cortical dynamics, a discrete selection can have continuously evolving effects on movement. Its consequences depend on the cortical state and sensory feedback, rather than specifying a fixed trajectory independently of the network. In the novel-movement experiment, sequencing individual adapters retained the movement's broad structure, while graded combinations improved trajectory accuracy (Figure~\ref{fig:transfer}; Supplementary Figure~\ref{fig:transfer-hard}). Selection and kinematic refinement may therefore be complementary uses of the same interface, CABRA being a concrete computational instantiation of that interface. 

An important direction for future work is to learn more of the structure currently supplied by the training setup. In our main experiments, two segments reflect the extension--return organization of the demonstrations, while the adapter-bank size sets the capacity of the available repertoire without prescribing task-to-adapter assignments. These choices influence the granularity of the learned operations: fewer adapters encourage sharing, whereas larger banks permit greater specialization but may also introduce redundant operations. Our reduced-capacity experiments retain low-dimensional adapter perturbations while increasing overlap between task representations, illustrating this tradeoff (Supplementary Figure~\ref{fig:adapter-capacity}). A natural extension is therefore to infer how many operations a behavioral repertoire requires, together with when those operations begin and end. The learning objectives also contain dataset-specific and engineering choices: task-identity supervision emphasizes information distinguishing the demonstrated movements, while identity and muscle-excitation objectives encourage adapter differentiation and accurate execution. Our Kitchen experiments demonstrate that CABRA can learn useful operations with less semantic supervision and task-specific structure: they omit task-identity supervision and the extension--return phase mask, instead learning from fixed-duration demonstration windows (Supplementary Figure~\ref{fig:kitchen-rl}). Learning useful decompositions with less semantic supervision and fewer structural assumptions is a valuable direction for future work to strengthen the generality of this architecture. 

The proposed mechanism suggests examining motor learning through the joint evolution of cortical dynamics and the signals that recruit them. Simultaneous cortical, thalamic, and basal ganglia recordings could test whether recurring control patterns accompany reproducible changes in cortical dynamics across movements. Perturbations could further distinguish disruption of an operation's execution from disruption of its selection, while neural-data constraints could help distinguish task-equivalent model solutions \citep{valente2022extracting}. Such experiments can use real data to test and ground the central implication of our theory, which is that learning a movement can establish more than the ability to reproduce it: it can establish an operation on shared neural circuitry that subsequent control can recruit for a different purpose.


\section*{Acknowledgments}
S.K is supported by a Leon Levy Fellowship by the New York Academy of Sciences.
\bibliographystyle{unsrtnat}
\bibliography{neurips_2026}

\appendix

\section{Supplementary Methods}
\label{appendix:methods}

\subsection{Architecture details}
\label{appendix:architecture}

CABRA uses one shared LSTM core in observation and execution modes, with separate input projections, selection heads, a muscle readout, and a bank of adapters. In the implementation, the observation pathway is termed the encoder and the execution pathway the decoder. The core and adapter hidden dimension is $H=256$.

\paragraph{Shared LSTM core.} A single LSTM of width $H=256$ is used both for encoder and decoder forward passes. Recurrent parameters are detached during the decoder pass, preventing direct execution-path updates to those weights. Hand-trajectory error still reaches the shared core indirectly through the encoder-produced control variables, as well as updating the decoder input projection, adapters, and muscle readout.

\paragraph{Input projections.} The encoder and decoder receive different observation subsets and use separate input projection MLPs (each a two-layer MLP with ReLU activation, hidden width $H{=}256$):
\begin{itemize}[leftmargin=1.2em,itemsep=0pt]
    \item Encoder input: 24-dim concatenation of the 18-dim sliced observation (full observation with the 10-dim rule input stripped) and the 6-dim teacher muscle command.
    \item Decoder input: 18-dim sliced observation only (no muscle command, no rule input).
\end{itemize}

\paragraph{Observation-mode selection and task heads.} Three heads read from the encoder LSTM hidden state. The boundary and code heads are two-layer MLPs ($H \to H \to \cdot$) with a ReLU between layers; the task head is also a two-layer MLP with ReLU.
\begin{itemize}[leftmargin=1.2em,itemsep=0pt]
    \item \emph{Boundary head}: outputs a scalar logit per timestep, softmaxed across time within each segment to yield $p(b)$.
    \item \emph{Code head}: outputs $K=10$ logits per segment. Sampled with the straight-through Gumbel-softmax estimator at temperature $\tau_z = 0.1$, yielding a one-hot $z_{m,k}$ in the forward pass with gradients flowing through the soft distribution.
    \item \emph{Task head} (auxiliary): two-layer MLP ($H \to 64 \to 5$) reading from the encoder's first-segment pooled hidden state, predicting task identity. Loss propagates only through the encoder.
\end{itemize}

\paragraph{Adapter bank.} Each of the $K=10$ adapters is a two-layer nonlinear residual modulator:
\begin{equation*}
    \Delta_k(\mathbf{h}) = \mathbf{W}_k^{\mathrm{out}} \tanh\!\left(\mathbf{W}_k^{\mathrm{in}} \mathbf{h}\right),
\end{equation*}
where $\mathbf{W}_k^{\mathrm{in}}, \mathbf{W}_k^{\mathrm{out}} \in \mathbb{R}^{H \times H}$. No bias terms. The output weights $\mathbf{W}_k^{\mathrm{out}}$ are scaled by a factor of $2$ at initialization to produce non-negligible early perturbations and accelerate adapter learning. No rank constraint is imposed on $\Delta_k$ in the architecture; the empirical rank of the learned adapter perturbation in the cortical hidden state is analyzed in Section~\ref{sec:results}.

\paragraph{Phase-split mask.} Codes are partitioned by movement phase: codes $z_0, \ldots, z_{K/2-1}$ are reserved for segment 0 (extension) and $z_{K/2}, \ldots, z_{K-1}$ for segment 1 (retraction). The phase split is implemented via hard masking of the invalid logits before sampling and is the only structural prior we impose on code identities; the assignment of tasks to specific codes within each phase emerges from training.

\paragraph{From segment selections to a control sequence.} For segment $m\in\{1,2\}$, let $z_{m,k}$ denote the sampled recruitment of adapter $k$, indexed from $0$ to $K-1$. If $\sigma(t)=\sum_{\tau\leq t}b_\tau$ is the cumulative probability that the inferred boundary has occurred, the time-dependent weights in Equation~\ref{eq:adapter-update} are $z_{t,k}=(1-\sigma(t))z_{1,k}+\sigma(t)z_{2,k}$. Thus discrete segment identities and a soft boundary produce the full control sequence; at evaluation, identities and boundary are hardened by argmax.

\paragraph{Muscle readout.} A linear layer maps the (adapter-perturbed) hidden state $\tilde{\mathbf{h}}_t$ to a 6-dimensional pre-activation, passed through a sigmoid to produce the muscle excitation vector supplied to the differentiable arm.

\paragraph{DIAYN discriminator.} The DIAYN \citep{eysenbach2018diversity} discriminator is implemented in \texttt{binned\_muscle} mode: muscle commands within each segment are pooled into $n_\text{bins}=3$ temporal bins and concatenated to form a $3 \times 6 = 18$-dim feature, fed to a two-layer MLP ($18 \to 64 \to K/2$) that predicts the active code restricted to the segment's valid phase subset. Separate discriminators are used for extension and retraction segments. The discriminator reads only muscle output, never cortical state.

\subsection{Training procedure}
\label{appendix:training}

\paragraph{Optimization.} Training uses Adam with learning rate $10^{-4}$, batch size $32$, gradient clipping at norm $1.0$, and no weight decay. The full system is trained end-to-end for $30{,}000$ iterations. Each training step pairs an observation pass and an execution pass through the shared LSTM.

\paragraph{Closed-loop rollouts.} At each training iteration, the encoder forward pass reads the full teacher demonstration (observation and muscle command) and produces $p(b)$, $p(z)$, and the task-identity logits. The decoder forward pass uses the sampled $z_{m,k}$ and $\sigma_m(t)$ to roll out the closed-loop arm trajectory using only body-state feedback. The hand-trajectory $L_1$ loss is computed between CABRA's closed-loop rollout at the current training step and the teacher's pre-recorded hand trajectory for the same trial. Both trajectories are produced by closed-loop rollouts through the same differentiable arm: the teacher's was recorded once during demonstration generation; CABRA's is produced fresh each training step from its current parameters.

\paragraph{Gradient paths between observation and execution.} The decoder uses detached copies of the shared recurrent parameters. Execution error cannot update those parameters through the execution recurrence, but can update them through the encoder that produced the sampled codes and boundaries. Detachment therefore separates gradient pathways; it does not insulate the shared core from all behavioral error signals.

\paragraph{Loss composition.} The total loss decomposes into three components corresponding to the encoder's segmentation/inference objective, the decoder's closed-loop control objective, and a set of auxiliary losses:
\begin{align*}
    \mathcal{L}_{\mathrm{enc}} &= \beta_z \mathcal{L}_{\mathrm{KL}_z} + \beta_b \mathcal{L}_{\mathrm{KL}_b} \\
    \mathcal{L}_{\mathrm{dec}} &= \mathcal{L}_{\mathrm{hand}} + \lambda_{\mathrm{rate}} \mathcal{L}_{\mathrm{rate}} + \lambda_{\mathrm{musc}} \mathcal{L}_{\mathrm{musc}} \\
    \mathcal{L}_{\mathrm{aux}} &= \beta_{\mathrm{DIAYN}} \mathcal{L}_{\mathrm{DIAYN}} + \lambda_{\mathrm{exc}} \mathcal{L}_{\mathrm{exc}} + \lambda_{\mathrm{task}} \mathcal{L}_{\mathrm{task}} \\
    \mathcal{L} &= \mathcal{L}_{\mathrm{enc}} + \mathcal{L}_{\mathrm{dec}} + \mathcal{L}_{\mathrm{aux}}
\end{align*}

The encoder objective $\mathcal{L}_{\mathrm{enc}}$ regularizes the boundary and code distributions toward priors: $\mathcal{L}_{\mathrm{KL}_z}$ is the KL between the encoder's per-segment code distribution and a uniform prior over the valid code subset; $\mathcal{L}_{\mathrm{KL}_b}$ is the KL between the boundary distribution and a per-sample Poisson prior centered on the trial midpoint.

The decoder objective $\mathcal{L}_{\mathrm{dec}}$ shapes the closed-loop motor output: $\mathcal{L}_{\mathrm{hand}}$ is the per-timestep $L_1$ between CABRA and teacher hand trajectories averaged over the trial; $\mathcal{L}_{\mathrm{rate}}$ and $\mathcal{L}_{\mathrm{musc}}$ are $L_1$ regularizers on decoder hidden activity and muscle excitations following \citet{lazzari2025multitasking}.

The auxiliary losses $\mathcal{L}_{\mathrm{aux}}$ provide additional learning signal: $\mathcal{L}_{\mathrm{DIAYN}}$ is the cross-entropy of the discriminator predicting the active code from segment-pooled muscle commands; $\mathcal{L}_{\mathrm{exc}}$ is an $L_1$ loss between CABRA and teacher muscle excitations; $\mathcal{L}_{\mathrm{task}}$ is the cross-entropy of the task-identity readout on the encoder's first-segment representation. The DIAYN and excitation-matching losses are ablated in Figure~\ref{fig:auxloss-ablation}; task-identity supervision remains in those motor ablations.

Loss weights for the headline configuration are $\beta_z = 0.3$, $\beta_b = 0.1$, $\beta_{\mathrm{DIAYN}} = 1.0$, $\lambda_{\mathrm{rate}} = 10^{-3}$, $\lambda_{\mathrm{musc}} = 10^{-2}$, $\lambda_{\mathrm{exc}} = 0.5$, $\lambda_{\mathrm{task}} = 0.5$.

Each training run was performed on a single NVIDIA GeForce GTX 1080 Ti GPU and took approximately 48 hours of wall-clock time to complete 30,000 iterations.

\subsection{Task and data details}
\label{appendix:tasks}

\paragraph{Biomechanical environment.} We use the \texttt{motornet} \citep{codol2024motornet} \texttt{RigidTendonArm26} effector with the \texttt{MujocoHillMuscle} model: a two-link arm with six muscles (two per joint plus two biarticular). Arm parameters match those used by \citet{lazzari2025multitasking}.

\paragraph{Compound tasks.} CABRA is trained on five compound trajectories (each comprising an extension followed by a retraction):
\begin{itemize}[leftmargin=1.2em,itemsep=0pt]
    \item \textbf{FullReach}: straight extension to target, then straight retraction home.
    \item \textbf{FullCircleClk}: clockwise half-circle outward, clockwise half-circle home.
    \item \textbf{FullCircleCClk}: counter-clockwise half-circle outward, counter-clockwise home.
    \item \textbf{Figure8}: half of a figure-8 outward, second half on return.
    \item \textbf{Figure8Inv}: figure-8 with inverted phase.
\end{itemize}
For each task we sample 32 reach conditions (target locations) per training batch. Each trial is a 200-timestep movement.

\paragraph{Demonstration information.} The encoder sees the teacher's realized muscle commands together with sensory observations, whereas CABRA in execution mode generates its own commands and receives its own body-state feedback. 

\paragraph{Half-task primitives are unseen.} Although each compound trajectory consists of an extension followed by a retraction, we never expose CABRA to the half-task primitives in isolation. Any decomposition into reusable extension and retraction codes must arise from segmenting compound demonstrations rather than from explicitly demonstrated primitives.

\subsection{Analysis methodology}
\label{appendix:analysis}

\paragraph{Effective rank.} For each adapter $k$, we collect the cortical perturbation $\Delta_k(\mathbf{h}_t)$ over a held-out task suite and compute the effective rank as the number of singular values needed to explain $\geq 95\%$ of the cumulative variance. Uncertainty is estimated from the standard deviation of rank estimates across $n_\text{boot}=10$ bootstrap resamples; individual-adapter estimates are then averaged. For this core-focused assay, the unmodulated core state is carried into the next recurrent update, while adapter-modulated output still drives the arm and its feedback. This differs from native execution, which carries the modulated state forward. The core, individual perturbations, and pooled perturbations are measured under this common assay.

\paragraph{Principal-subspace overlap.} For each task $A$, we collect cortical hidden state activity over the trajectory into $\mathbf{X}_A \in \mathbb{R}^{N \times T}$ ($N=256$, $T$ timesteps) and extract the top $m$ principal components $\mathbf{P}_A \in \mathbb{R}^{N \times m}$ explaining $\geq 95\%$ of task variance. We use $m_{\mathrm{teacher}}=12$ and $m_{\mathrm{CABRA}}=25$, each matching its network's effective rank under the 95\% variance threshold. The pairwise principal-subspace overlap between tasks $A$ and $B$ is
\begin{equation*}
    \mathrm{overlap}(A, B) = \frac{1}{m} \sum_{i=1}^{m} \cos^2 \theta_i, \quad \{\cos \theta_i\}_{i=1}^{m} = \mathrm{svd}\!\left(\mathbf{P}_A^\top \mathbf{P}_B\right),
\end{equation*}
where $\theta_i$ are the principal angles between the two subspaces. The chance baseline is the expected overlap between two random orthonormal $m$-dimensional subspaces in $\mathbb{R}^N$, computed via QR decomposition of Gaussian matrices ($\sim 0.05$ for $m=12$, $\sim 0.10$ for $m=25$). Trajectories used for this analysis are pooled across $128$ trials per task ($32$ reach conditions $\times 4$ repeats).

\paragraph{Dynamical Similarity Analysis.} We follow the protocol of \citet{lazzari2025multitasking}, computing DSA \citep{ostrow2023beyond} pairwise between tasks. For each task we collect cortical activity into a tensor over conditions, trials, timesteps, and units, PCA-reduce to the network's effective rank ($m_\text{teacher}=12$, $m_\text{CABRA}=25$), construct delay-embedded Hankel tensors with lag $p=90$, and fit reduced-rank Dynamic Mode Decomposition (DMD) matrices $\mathbf{A}_A, \mathbf{A}_B$ via Hankel Alternative View of Koopman (HAVOK) at rank $r=150$. The DSA distance between tasks is the Procrustes distance under similarity transformation:
\begin{equation*}
    d(\mathbf{A}_A, \mathbf{A}_B) = \min_{\mathbf{C} \in O(r)} \|\mathbf{A}_A - \mathbf{C}\, \mathbf{A}_B\, \mathbf{C}^{-1}\|_2,
\end{equation*}
computed using the \texttt{DSA} package with \texttt{score\_method="euclidean"}. All DSA hyperparameters match \citet{lazzari2025multitasking}.

\paragraph{Transfer optimization.} For both teacher and CABRA, we freeze the network and optimize a $K$-dimensional soft policy over the network's task interface (rule-input weights for the teacher, adapter-code weights for CABRA). Each policy is a learnable matrix of logits over time, softmaxed at temperature $\tau = 0.5$ to produce a per-timestep distribution over the $K=10$ task interfaces. We optimize hand-trajectory $L_1$ loss against the two-petal target using Adam with learning rate $0.1$ for $350$ iterations. Both networks use identical optimizer settings, target trajectory, and policy initialization; the only difference is the frozen substrate. The butterfly trajectory uses $600$ iterations at the same learning rate to accommodate its more complex shape. The softmax temperature in principle allows either policy to mix multiple task interfaces simultaneously at any timestep; differences in how soft each network's optimized policy ends up therefore reflect each network's underlying representational structure rather than any architectural asymmetry in the optimization itself.

\paragraph{Discrete selection and mixing.} Code sampling, boundary sampling, and transfer optimization have distinct roles. Motor training uses straight-through discrete codes and a soft boundary assignment; evaluation hardens the inferred program. Novel-trajectory optimization instead permits graded adapter mixtures throughout the movement, not only at boundaries. The hard-selection control uses straight-through one-hot execution under the same transfer objective (Figure~\ref{fig:transfer-hard}); it retains a transfer advantage at reduced accuracy. We did not perform a temperature sweep.

\paragraph{Forced-code analysis.} For the qualitative inspection of code-task assignment (Figure~\ref{fig:behavior}), we run CABRA in closed loop while overriding the encoder's code distribution with a fixed one-hot code per segment. The resulting hand trajectory shows what behavior each adapter $\Delta_k$ produces in isolation, factored away from the encoder's learned code-inference policy.

\subsection{Kitchen skill learning and reinforcement learning}
\label{appendix:kitchen}
We used demonstration trajectories from the Kitchen manipulation environment with 60-dimensional state observations and nine-dimensional actions. CABRA learned to predict actions over ten-step demonstration windows using a shared recurrent core of width 256 and ten full-width residual adapters. The demonstration encoder inferred a categorical skill distribution, and the decoder predicted actions conditioned on observed states and the inferred skill. Unlike the differentiable MotorNet rollout objective, this stage used action reconstruction on demonstration states. No task-identity classifier or two-phase mask was used, and window duration was fixed rather than inferred.

The offline objective combined action mean-squared error with aggregate code-distribution KL regularization toward a uniform distribution (weight $0.3$), posterior-entropy regularization ($0.01$), a state-conditioned prior-matching loss ($0.1$), a skill-discrimination objective ($1$), and adapter-output and core-activity penalties ($10^{-5}$ each). The prior predicted the inferred skill from the window's initial state, with the encoder posterior detached in the prior-matching target. Training used Adam at learning rate $10^{-4}$, batch size 128, for 100 epochs.

For downstream control, we followed the skill-prior-regularized hierarchical reinforcement-learning setup of SPiRL \citep{pertsch2021accelerating}. Low-level networks remained frozen. A high-level policy selected a skill every ten environment steps and learned from task reward with KL regularization toward the frozen state-conditioned prior. CABRA used a categorical policy over adapters, initialized from the learned prior with a zero-initialized residual policy, and a discrete soft actor-critic update with exact categorical KL and target divergence $0.5$. The SPiRL comparison used its continuous latent-skill policy and corresponding skill prior. Thus both methods use prior-regularized hierarchical control, but differ in skill parameterization and the associated policy implementation. Figure~\ref{fig:kitchen-rl} shows the first 5,000 training episodes. Episode reward counts completed subtasks, up to four. Curves use nonoverlapping 100-episode bins and bands are the within-bin standard errors. 


\section{Supplementary Figures}
\label{appendix:supp-figures}

\begin{figure}[h]
    \centering
    \includegraphics[width=0.8\textwidth]{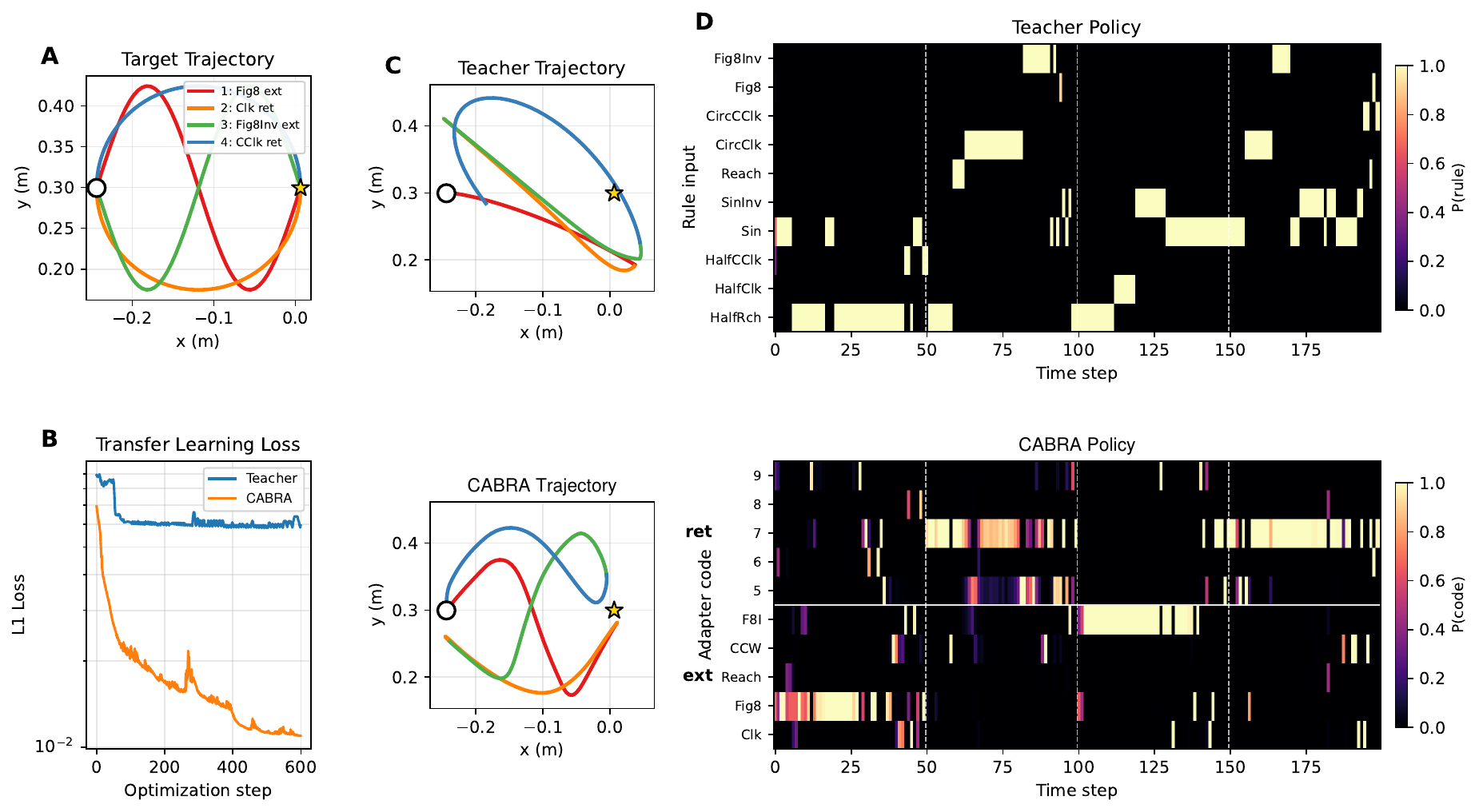}
    \caption{\textbf{Compositional transfer to a butterfly trajectory.} A second out-of-distribution test using a four-segment butterfly composed of (Figure-8 extension, Clockwise retraction, Figure-8 Inverse extension, Counter-clockwise retraction); none of these pairings appeared in either network's training data.
    \emph{(A)} The butterfly target trajectory, with each 50-timestep segment shown in a separate color. Open circle marks the home position; star marks the reach target.
    \emph{(B)} Hand $L_1$ during transfer optimization. Both networks are frozen and a 10-dimensional soft policy over the network's task interface is optimized for 600 steps. CABRA reaches hand $L_1 \approx 0.011$, while the teacher plateaus at $\approx 0.058$, a $\sim 5\times$ gap.
    \emph{(C)} Final teacher (top) and CABRA (bottom) hand trajectories. CABRA recovers the butterfly's figure-8-like crossing structure; the teacher's trajectory is qualitatively distorted.
    \emph{(D)} Final policies. Top: teacher rule-input weights over time. Bottom: CABRA adapter-code weights.}
    \label{fig:butterfly}
\end{figure}

\begin{figure}[h]
    \centering
    \includegraphics[width=0.8\textwidth]{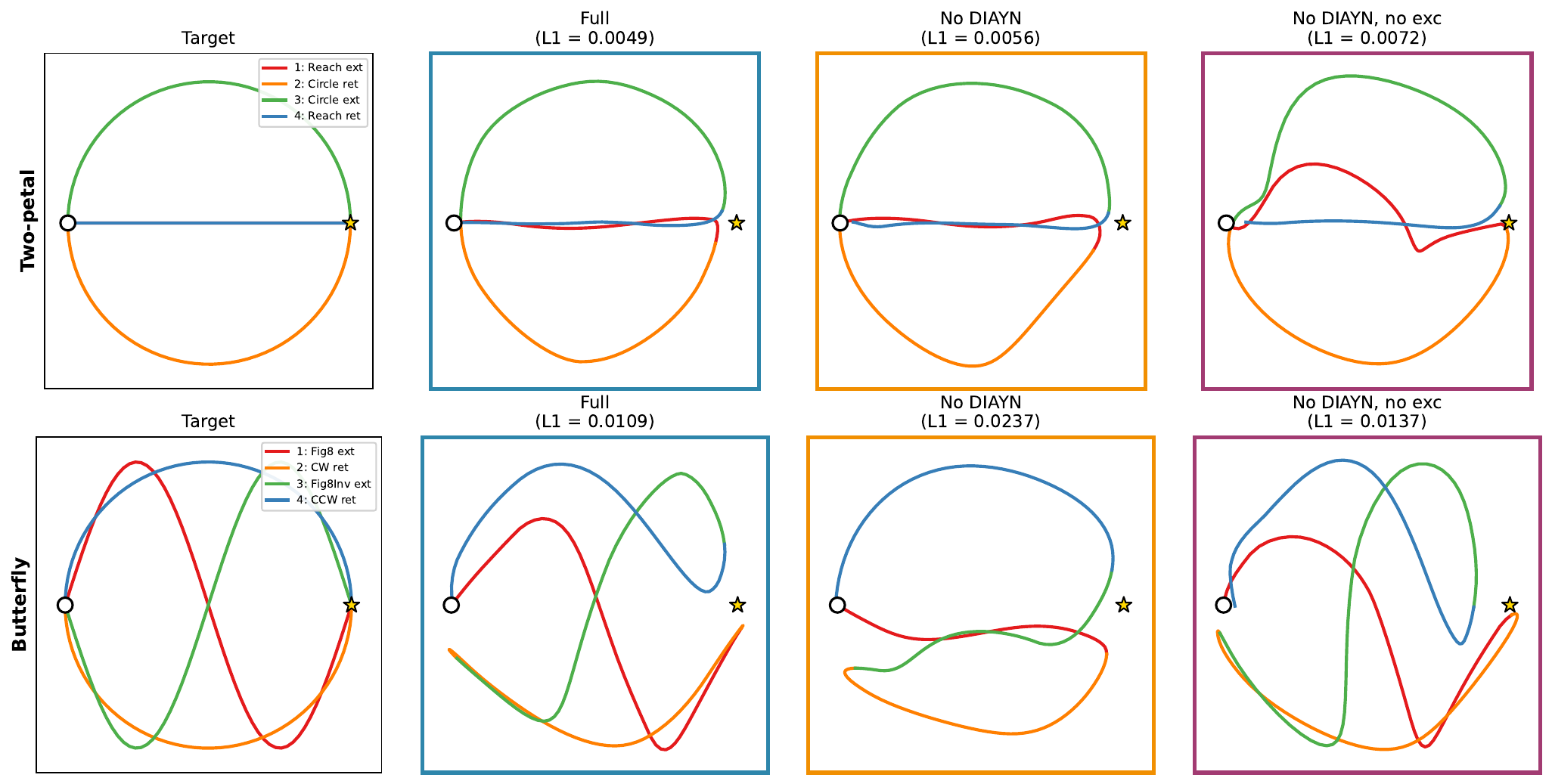}
    \caption{\textbf{Ablation of auxiliary supervision components across two OOD trajectories.} We test whether the architectural transfer advantage depends on the two auxiliary losses ($\mathcal{L}_{\mathrm{DIAYN}}$ and $\mathcal{L}_{\mathrm{exc}}$) by training two ablated configurations: one with $\beta_{\mathrm{DIAYN}} = 0$ (No DIAYN), and one with both $\beta_{\mathrm{DIAYN}} = 0$ and $\lambda_{\mathrm{exc}} = 0$ (No DIAYN, no exc). Both ablations retain the adapter architecture, phase mask, and task-identity objective. Final transfer trajectories on the two-petal (top) and butterfly (bottom) OOD tasks across the target and three configurations. The full configuration achieves numerically the best transfer hand $L_1$ on both OOD trajectories (two-petal: $0.005$; butterfly: $0.011$), and all three configurations achieve transfer well below the teacher baseline (two-petal: $0.04$, butterfly: $0.058$). The transfer advantage persists without DIAYN and excitation matching; task-identity supervision remains in all three conditions.}
    \label{fig:auxloss-ablation}
\end{figure}

\begin{figure}[t!]
    \centering
    \includegraphics[width=0.78\textwidth]{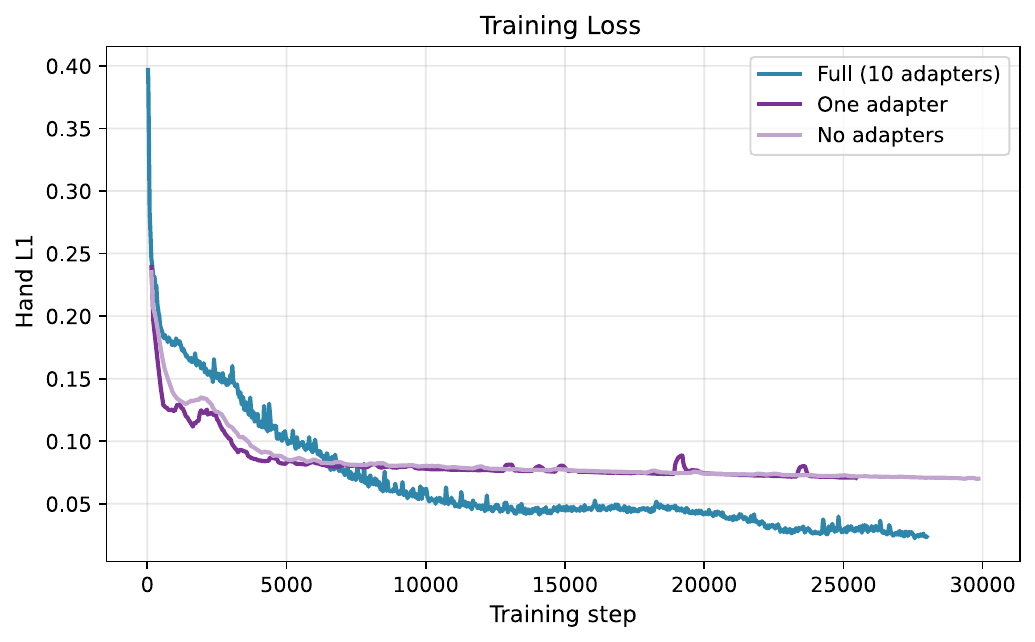}
    \caption{\textbf{Adapter bank ablation.} The previous ablation removed auxiliary supervision components but preserved the adapter architecture. We additionally test whether the adapter bank itself is crucial for the network's ability to learn the multi-task suite by training two further ablations: one with a single shared adapter ($K{=}1$, with all tasks routed through the same modulator), and one with no adapters at all (the cortex must learn all task dynamics within a single shared recurrence). For the latter, we allowed the decoder loss to train the LSTM recurrence.}
    \label{fig:adapter-ablation}
\end{figure}

\begin{figure}[h]
\centering
\includegraphics[width=\textwidth]{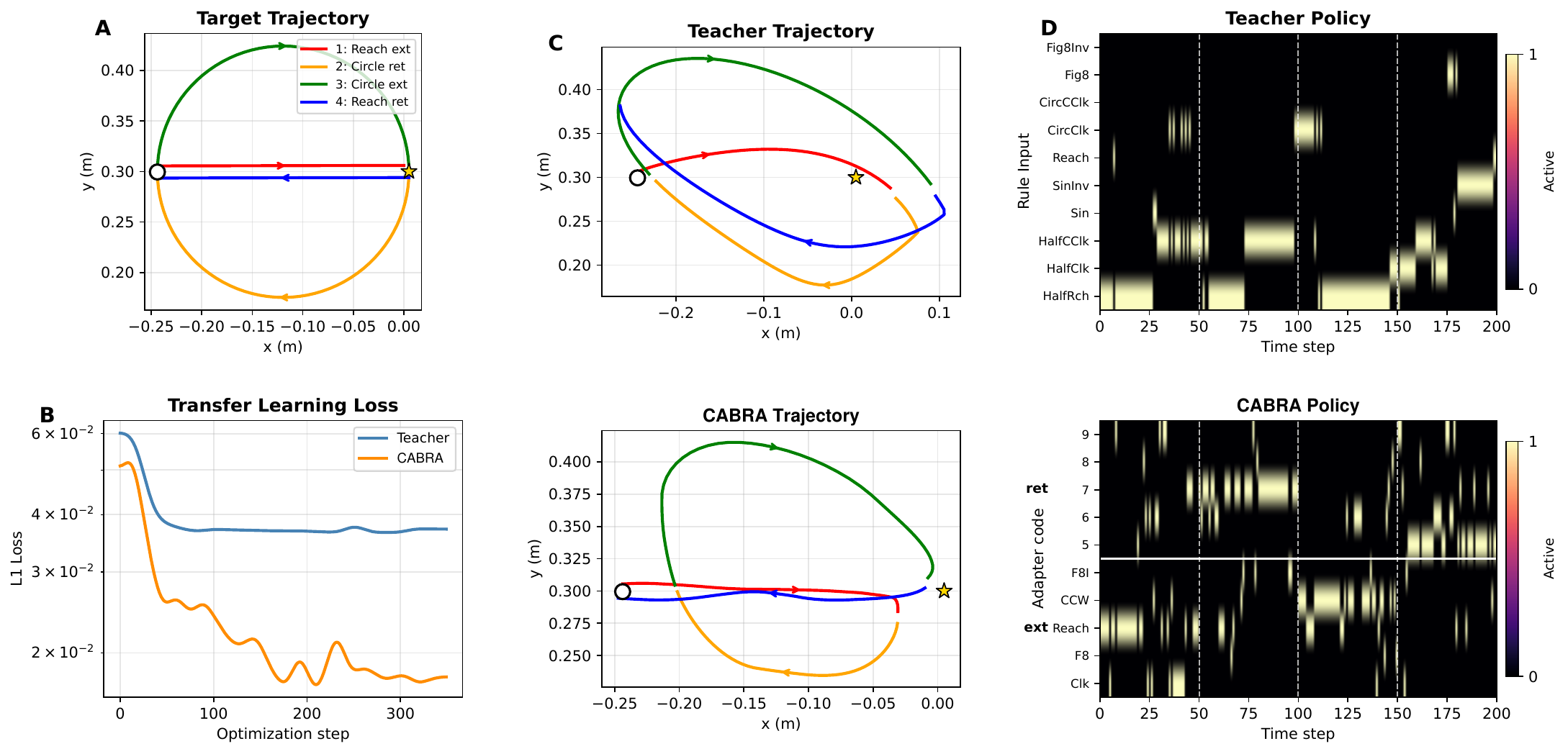}
\caption{\textbf{Compositional transfer under hard discrete policy execution.}
A complementary version of the two-petal transfer experiment in
Figure~\ref{fig:transfer}, in which both networks are constrained to use
\emph{hard one-hot} selection over their respective task interfaces at every
timestep. Optimization uses a straight-through estimator: the forward pass
takes the argmax of the policy logits at each timestep (genuinely discrete
selection), while gradients flow through the soft softmax distribution
during backpropagation.
\textbf{(A)} The two-petal target trajectory (same as Figure~\ref{fig:transfer}).
\textbf{(B)} Smoothed L1 loss over 350 optimization steps. Both networks are
optimized with identical Adam settings, target trajectory, and initialization
heuristic; they differ only in the frozen substrate. Curves show
median-filtered loss (window 25) followed by Gaussian smoothing ($\sigma=8$)
to suppress the noise inherent to straight-through estimation. Final mean L1
over the last 50 steps: CABRA $0.018$, teacher $0.040$, a $2.2\times$ gap;
best L1: CABRA $0.014$, teacher $0.036$, a $2.5\times$ gap.
\textbf{(C)} Final teacher (top) and CABRA (bottom) hand trajectories. Even
under hard discrete selection, CABRA recovers the qualitative two-petal
structure (orange and green half-loops), while the teacher's trajectory is
contracted and distorted.
\textbf{(D)} Final policies as actually applied to each network: the argmax
code at each timestep. Top: teacher rule-input selection; bottom: CABRA
adapter-code selection.}
\label{fig:transfer-hard}
\end{figure}

\clearpage
\begin{figure}[p]
\centering
\includegraphics[width=\linewidth]{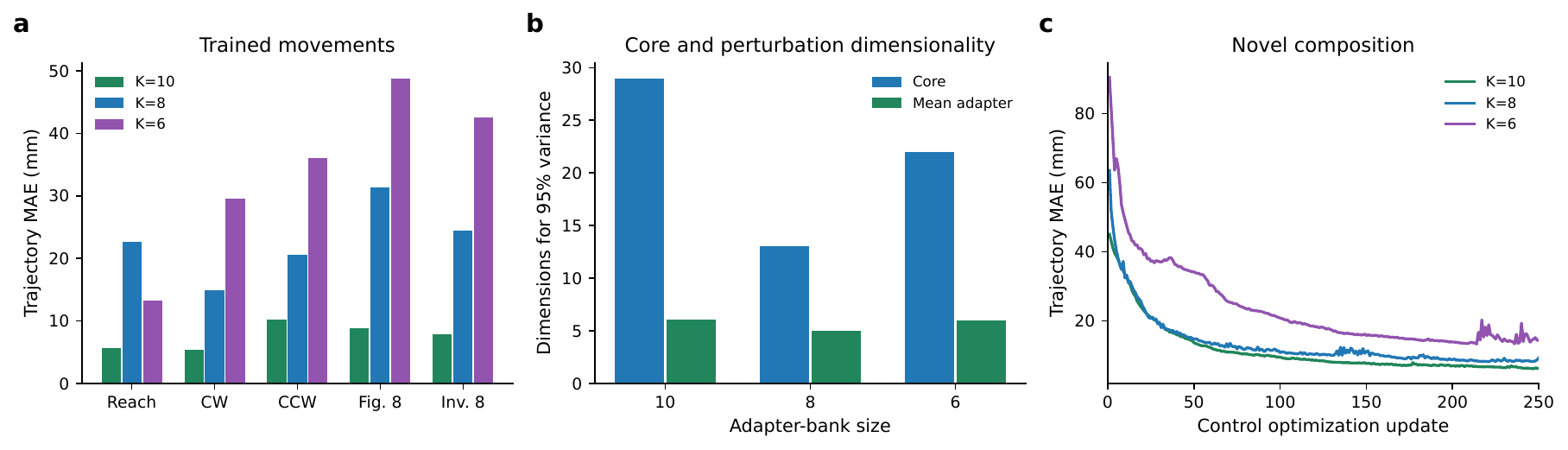}
\caption{\textbf{Smaller adapter banks retain low-dimensional modulation and compositional transfer.}
\textbf{(a)} Bars show mean trajectory error on the five trained movements for each adapter-bank size, evaluated on 32 demonstrations per movement.
\textbf{(b)} Dimensionality of core activity and mean individual adapter perturbations, measured by the number of dimensions explaining 95\% of variance using the core-focused rank protocol.
\textbf{(c)} Optimization of new control programs for the two-petal trajectory over frozen networks with 10, 8 or 6 adapters, shown through update 250.
Smaller banks preserve the ability to compose a new movement, with reduced accuracy in these runs.}
\label{fig:adapter-capacity}
\end{figure}

\begin{figure}[p]
\centering
\includegraphics[width=\linewidth]{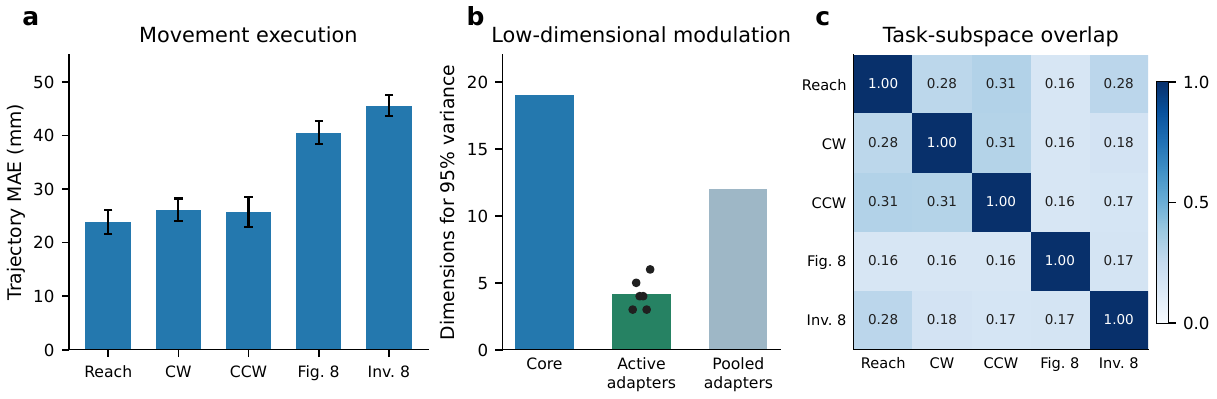}
\caption{\textbf{Low-dimensional adapter perturbations and separated task representations emerge without a phase-split mask.}
All ten adapters were available during both extension and retraction, retaining two inferred segments per movement.
\textbf{(a)} Trajectory MAE on trained movements, averaged across time and both spatial coordinates; bars and error bars show mean $\pm$ SEM over 32 demonstrations per movement.
\textbf{(b)} Core activity occupies 19 dimensions, individual active-adapter perturbations occupy 4.17 dimensions on average, and pooled perturbations occupy 12 dimensions (95\% variance criterion). Dots show the six adapters engaged by the analysis programs, and the green bar shows their mean. The other four adapters were not recruited in these movements and thus left out. 
\textbf{(c)} Pairwise overlap of post-adapter task subspaces, measured as the mean squared cosine of principal angles using 25 principal components. Mean off-diagonal overlap is 0.219. Geometry was measured from closed-loop execution of the dominant inferred adapter pair for each task, with a midpoint transition.}
\label{fig:no-phase-mask}
\end{figure}
\begin{figure}[p]
\centering
\includegraphics[width=\linewidth]{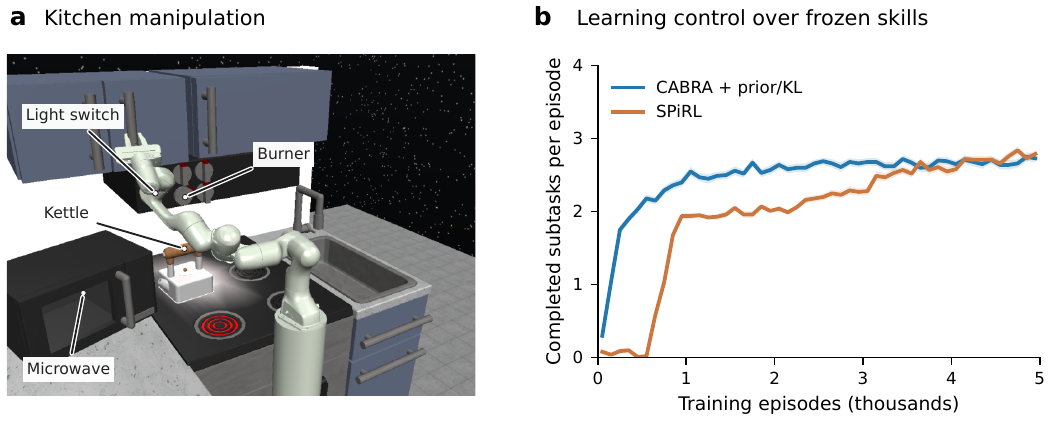}
\caption{\textbf{CABRA supports efficient reinforcement learning in Kitchen.}
\textbf{(a)} A simulated 3D robotic arm manipulates a microwave, kettle, burner and light switch.
\textbf{(b)} After demonstration pretraining, a high-level policy learns to select skills over a frozen low-level controller. Both methods use a demonstration-derived skill prior and KL-regularized high-level reinforcement learning; CABRA uses a discrete adapter bank in place of SPiRL's latent skill interface. CABRA pretraining here uses no task-identity supervision. Curves show the mean number of completed subtasks in nonoverlapping blocks of 100 training episodes, plotted at block centers, for one run per method. Shading shows $\pm$1 SEM within each 100-episode block. CABRA acquires rewarded behavior earlier and reaches similar late performance. Mean reward over last 200 episodes is 2.74 for CABRA and 2.76 for SPiRL (maximum: four subtasks).}
\label{fig:kitchen-rl}
\end{figure}

\clearpage 

\end{document}